\documentclass{llncs}
\usepackage{amsfonts}
\usepackage[cmex10]{amsmath}
\usepackage{graphicx}
\graphicspath{{./figures/}}
\usepackage[tight,footnotesize]{subfigure}
\usepackage{makeidx}  

\usepackage{multirow}
\usepackage{tabularx}

\usepackage{tikz}
\usetikzlibrary{shapes.geometric,arrows}

\tikzstyle{startstop} = [rectangle, rounded corners, minimum width=3cm, minimum height=0.25cm, text centered, draw=black, fill=red!30]
\tikzstyle{io} = [trapezium, trapezium left angle=70, trapezium right angle=110, minimum width=3cm, minimum height=0.5cm, text centered, draw=black, fill=blue!30]
\tikzstyle{process} = [rectangle, minimum width=3cm, minimum height=0.25cm, text centered, draw=black, fill=orange!30]
\tikzstyle{decision} = [diamond, minimum width=0.25cm, minimum height=0.25cm, text width=1.5cm, text centered,  draw=black, fill=green!30] 
\tikzstyle{arrow} = [thick,->,>=stealth] 

\begin{document}
\title{Optimization of OFDM radar waveforms using genetic algorithms}
\titlerunning{Multi-carrier: GA}  
%
\author{Gabriel Lellouch
 \and Amit Kumar Mishra }
\authorrunning{Lellouch et al.} 
%
\tocauthor{Gabriel Lellouch and Amit Kumar Mishra}
\institute{University of Cape Town, South Africa,\\
\email{gabriel.lellouch@gmail.com\\akmishra@ieee.org},
}

\maketitle              

\begin{abstract}

\keywords{OFDM Radar, Genetic Algorithm, NSGA-II, PSLR, ISLR, PMEPR}
\end{abstract}

In this paper, we present our investigations on the use of single objective and multiobjective genetic algorithms based optimisation algorithms  to improve the design of OFDM pulses for radar. 
We discuss these optimization procedures in the scope of a waveform design intended for two different radar processing solutions. 
Lastly, we show how the encoding solution is suited to permit the optimizations of  waveform for OFDM radar related challenges such as enhanced detection.
\section{Introduction}
With the advent of powerful digital hardware, software defined radio and radar have become an active area of research and development \cite{Langman13}. This in turn has given rise to many new research directions in the radar community, which were previously not comprehensible. One such direction is the recently investigated OFDM radar \cite{Lellouch14b}, radars which use OFDM waveforms instead of the classic linear frequency modulated (LFM) waveform. 

OFDM is a special form of multicarrier modulation (MCM), where a single data stream is transmitted over a number of lower rate subcarriers. 
Alphabets such as BPSK, QPSK, etc., are commonly used to code the information. The resulting complex symbols, also called phase codes (-1,+1 in BPSK) are modulate the subcarriers \cite{Hara03}. 
When  signal gets to the receiver, a demodulation stage retrieves the transmitted phase codes and eventually the binary message. In radar, the priority is to detect the presence of targets and possibly estimate some of their features through the following measurable quantities: range, Doppler, azimuth and elevation. It is thus of utmost interest for the radar designer to understand the possibilities offered by the OFDM structure and optimize it to suit its needs.

It can be quickly verified that non-coded OFDM pulses will not be suitable in radar systems that operate with the conventional matched filter processing, since they will give rise to high sidelobes. 
Another drawback of non-coded OFDM pulses is their varying time domain signal. Strong variations are detrimental since these may saturate the signal and cause distortions. Therefore, the OFDM pulse needs to be tailored before it becomes a suitable radar waveform.

In this work, we argue that the emerging evolutionary algorithms are particularly adapted to solve the pulse design problem when the pulse is an OFDM signal. Although a number of techniques have been proposed to mitigate the peak to mean envelope power ratio (PMEPR) and the peak-to-sidelobe ratio (PSLR) \cite{Levanon04}, their flexibility is somewhat limited. For example, the Newman phasing technique gives very low PMEPR for the single OFDM symbol case, and this for most numbers of subcarriers $N$. However, as soon as some subcarriers are suppressed the PMEPR deteriorates.

It may be remarked here that it is not the scope of this paper to investigate the use of newer evolutionary techniques, rather we show how we can integrate some simple and easy-to-implement memetic computing techniques in the design  of waveforms for  OFDM  radar. 
We will focus firstly on the genetic algorithm (GA) optimization technique and then on the multiple objective optimization genetic algorithm (MOO-GA) based technique. Although the former method offers a straightforward implementation \cite{Whitley}, many implementations exist in the case of the MOO-GA. In this work, we use the well known non-dominated sorting genetic algorithm II (NSGA-II). It has proved to be much faster than the earlier version NSGA while providing diversity in the solutions \cite{Deb02}.  

There are two major novelties of this work. 
First of all, the use of OFDM pulses as a radar signal is in itself a new direction. Secondly, the use of GA based techniques to design OFDM radar pulses is also novel. The rest of the paper is organized as follows. In section~\ref{sec:sect2} we present the first step of our waveform design and characterize the impact of the processing solution on this design. The objective here is to fix some of the OFDM parameters to build the frame of the waveform. We also stress the interconnection of the different OFDM parameters. In section~\ref{sec:sect3} we discuss the details of the optimization. We review the various objective functions, discuss the parameters that we use in our optimization problem and give our motivations to use GA based techniques as compared to other existing methods. In section~\ref{sec:sect4}, we review both of our GA based optimization methods. We present the technical details of the encoding strategy as well as the population size. We then present and discuss  our results in section~\ref{sec:sect5}. We start with the case of single objective optimization before moving on to the multiple objective optimization case. In section~\ref{sec:sect6}, we propose a case study where the single objective GA is integrated into a waveform design procedure for enhancing the target detection. We give our conclusions in section~\ref{sec:sect5}.

\section{Waveform design}\label{sec:sect2}
In this section, we present the successive steps that form our design strategy for characterizing the fixed parameters of the pulses. We show how they are inferred from, on the one hand, the processing and on the other hand, the scenario. This analysis fixes the frame of our pulses. We will show in the following section how we optimize the rest of our free parameters to compose pulses with improved radar features.  

\subsection{Processing related constraints}\label{sec:sect2a}
In the scope of a pulsed OFDM radar waveform, we proposed in \cite{Lellouch14b} two processing alternatives. The first alternative is based on the combination of matched filtering and Doppler processing whilst the second alternative transforms the received signal in the frequency domain in the same way as OFDM communication systems operate. After a demodulation stage which suppresses the phase codes, two orthogonal DFT processing are applied to form a range Doppler image. The key characteristics of both processing are recalled in Table~\ref{tab:processing_charact}. In the rest of the paper we refer to the former as the conventional processing while we name the latter our frequency domain processing. 

\begin{table}[h]
\caption{Conventional processing characteristics} 
\centering 
\setlength{\extrarowheight}{1.5pt}
\begin{tabular}{c|p{4cm}|p{4.9cm}}
 & \textbf{Conventional processing} & \textbf{Frequency domain processing} \\\hline
 \multirow{2}{*}{\textbf{Pros}}&Immune to intersymbol interference & Range and Doppler sidelobes are phase codes independent\\
    &Doppler sidelobes are phase codes independent \\
    \hline
\textbf{Cons}&Range sidelobes are phase codes dependent &Subject to intersymbol interference\\ 
\end{tabular}
\label{tab:processing_charact}
\end{table}

Because the frequency domain processing is subject to inter-symbol interference, we will use it to track targets. Indeed, in tracking configurations we can assume to have some prior knowledge of the illuminated scene and in particular the target extent. The rule of thumb is that the return echo from the closest point scatterer and the return echo from the furthest point scatterer of the target fall within the same time cell. When satisfied, this condition insures that the orthogonality between the subcarriers is maintained. This issue has been of utmost interest in the early years of OFDM signalling for communication to cope with the multipath effect. To that end the concept of cyclic prefix has been introduced \cite{Hara03}. In our current analysis, rather than inserting a cyclic prefix, we choose to match the size of the time cell according to the target extent. Since the conventional processing does not come up with a severe design constraint we choose to use the constraint of the frequency domain processing as the main guideline.

\subsubsection{Sampling frequency}\label{sec:sect2a1}
In our analysis, the received signal which we feed into either of these processing is the complex signal formed from the real and imaginary components respectively in the I and Q channels of the receiver. If the transmitted OFDM pulse has a bandwidth $B$, the received complex signal has the same bandwidth. Because the signal is complex, the Nyquist theorem states that the sampling frequency can be taken as low as $f_s=B$ and the time cell size is thus inversely proportional to the bandwidth, $t_s=1/B$. The size of the range cell is then given by $c/2B$, where $c$ is the speed of light. 

\subsubsection{Bandwidth}\label{sec:sect2a2}
As a result, if we want to design the size of the time cells such that, despite the superposition of all echoes returning from the different point scatterers of the target, the orthogonality property of the subcarriers is maintained, we shall adjust the bandwidth to comply with: $c/2B\geq \Delta R_t$ where $\Delta R_t$ is the target range extent. Practically we can add a margin to account for the target radial velocity and the uncertainty on the target extent and position. Not only the received echoes shall remain in the same time cell as a result of the first pulse but also throughout the coherent processing interval. In the end the bandwidth can be obtained from:
\begin{equation} \label{eq:Bandwidth}  
B = \frac{c}{2(\Delta R+\alpha)}
\end{equation}
where $\alpha$ is the margin in range.

Therefore, if there is no need for high range resolution we suggest to base the bandwidth selection on intersymbol interference mitigation considerations instead. When the radar is in a tracking mode, the target is known and the use of high resolution range profiles is a fortiori not necessary.

\subsection{Scenarios related constraints}\label{sec:section2b}
Other parameters need to be fixed. The pulse length and the number of subcarriers that will compose the OFDM signal.

\subsubsection{Pulse length}\label{sec:section2b1}
Pulse compression waveforms have the unique advantage to offer low peak power transmissions over a long time to provide the same maximum detection range and the same range resolution as would be obtained from a short pulse with a high peak power. But since the radar receiver is switched off when the pulse is being transmitted we cannot afford to have a very long pulse. We commonly refer to as the eclipsed zone, the window that lies between the radar and the minimum detection range. Even though we could choose one value for each target, for simplicity we choose only one for all. If, say, we expect targets from $R_{\text{min}}=1.5$ km, an upper bound for the pulse length is found to be \cite{Richards10}, $t_p=2\cdot R_{\text{min}}/c=10$ $\mu$s. 

\subsubsection{Number of subcarriers}\label{sec:section2b2}
The orthogonality property is another example of the unique OFDM structure. It states that the bit duration $t_b$ is inversely proportional to the subcarrier spacing $\Delta f$, $t_b=1/\Delta f$. In the extreme case where the pulse is composed of one symbol the maximum number of subcarriers $N_{\text{max}}$ used in the pulse can be derived from: 
\begin{equation} \label{eq:Nmax}  
N_{\text{max}} = \frac{2BR_{\text{min}}}{c}
\end{equation}
For the same pulse bandwidth and pulse duration a smaller number of subcarriers can be used if we construct the pulse from several symbols. For example we can decide to use 250 subcarriers and have 4 symbols in the pulse to maintain the same duration. The subcarrier spacing is then increased from 100 kHz to 400 kHz.  

In light of the above analysis, the fixed parameters for our pulses are summarized in Table~\ref{tab:scenar_charact}.

\begin{table}[!ht]
\caption{Scenarii characteristics for the waveform design} 
\centering 
\begin{tabular}{l | cc | cc } 
 && Case 1 && Case 2 \\ [0.5ex]
 && (walker) && (truck) \\ [0.5ex] 
\hline  
Range extent (m)&& 2 && 10 \\ 
Margin (m) && 1 && 5 \\
\hline 
\textbf{Bandwidth (MHz)}&& \textbf{50} && \textbf{10} \\ 
\textbf{Maximum number of subcarriers}&& \textbf{500} && \textbf{100} \\ 
\end{tabular}
\label{tab:scenar_charact} 
\end{table}

\section{Optimizing the pulse for radar}\label{sec:sect3}

Having fixed the frame of our pulses, literally the bandwidth and the maximum number of subcarriers, we are now ready to concentrate our analysis on the optimization of the OFDM pulse for radar. Firstly, we need to define our objective functions and secondly, we need to identify the OFDM parameters that we will use to run our optimization procedure.  

\subsection{Objective functions}\label{sec:sect3a}
An OFDM symbol is built as a sum of weighted complex sinusoids, where every sinusoid has a given starting phase. When the OFDM pulse is composed of several symbols it can be expressed as:
\begin{equation} \label{eq:OFDMbaseband}  
x(t) = \sum_{n=1}^{N} \sum_{k=1}^{K} w_n a_{n,k} r_k(t)\cdot \exp(j2\pi n\Delta f t).
\end{equation}
$N$ is the number of subcarriers, $w_n$ corresponds to the weight applied on subcarrier $n$ and the phase code $a_{n,k}$ is attributed to subcarrier $n$ in symbol $k$. $K$ is the total number of symbols in the pulse. The function $r_k(t)$ refers to the rectangular window for every symbol: 
\[ r_k(t) = \left\{
  \begin{array}{l l}
    1 & \quad (k-1)t_b \leq t\leq kt_b\\
    0 & \quad \text{elsewhere}
  \end{array} \right.\]

When using OFDM for radar two important aspects must be considered.
 
\subsubsection{Sidelobe level}\label{sec:sect3a1}
Firstly, would the processing be based on a correlation function like in our conventional processing where matched filtering is applied in range, the sidelobes at the output shall be maintained as low as possible. This consideration is true for any signal and a fortiori for our OFDM signal. 
When both functions are equal, the correlation function becomes the autocorrelation function. The output of $R(\tau)$ is then given by:
\begin{equation} \label{eq:Xcorr_cont} 
R(\tau) = \int_{-\infty}^{\infty}x(t)x^*(t-\tau)dt.  
\end{equation} 
In our analysis, we rather use the discrete form $R[m]$:
\begin{equation} \label{eq:Xcorr_disc} 
R[m] = \sum_{p=0}^{NK-1}x[p]x^*[p-m],  
\end{equation} 
where $m$ takes integer values between $-NK+1$ and $NK-1$. $x[p]$ represent the discrete values of the OFDM pulse taken at the discrete instants $pt_b/N=$, where $p$ takes integer numbers from 0 to $NK-1$. 
If the pulse is composed of only one symbol, then $p$ takes values from 0 to $N-1$, just like $n$ the subcarrier index. In the end, $x[p]=x(pt_b/N)$. In Eq.~\ref{eq:Xcorr_disc}, we assume that $x[p]=0$ for all forbidden values of $p$, that is $p<0$ and $p>NK-1$.

To cope with practical applications we commonly distinguish two objective functions. The first function is the peak sidelobe level ratio (PSLR). It returns the ratio between the highest sidelobe and the peak.
\begin{equation} \label{eq:PSLR} 
PSLR = \frac{\underset{m}{\operatorname{max}}  |R[m]|}{|R[0]|}, m\neq 0  
\end{equation} 
The second function is the integrated sidelobe level ratio (ISLR). It returns the ratio between the cumulation of the sidelobes and the peak. 
\begin{equation} \label{eq:ISLR} 
ISLR = \frac{\underset{m}\sum |R[m]|}{|R[0]|}, m\neq 0 
\end{equation}
The relative importance between these two figures of merit depends on the application as well as the environment. For example, if the radar operates in the presence of distributed clutter, it will be important to work with low ISLR in order to keep the weak targets visible. In that case, high ISLR can be interpreted as an increase of the noise floor. Conversely if the application requires detection of targets in the presence of strong discrete clutter, the PSLR is more critical and must be kept low to prevent from deceptively considering one sidelobe as another small target.

\subsubsection{Peak to mean envelope power ratio}\label{sec:sect3a2}
This second aspect is specific to OFDM signals. Its impact on the radar performance is less straightforward, nonetheless it is as relevant as the previous PSLR and ISLR. It characterizes the variations in time of the envelope. Too strong variations can be detrimental to the radar since the signal may saturate. A common expression for this peak-to-mean envelope power ratio (PMEPR) is given by \cite{Jones94}:
\begin{equation} \label{eq:PMEPR} 
PMEPR = \frac{\underset{n}{\operatorname{max}} |x[n]|^2}{\frac{1}{N}\sum|x[n]|^2},  
\end{equation}

\begin{figure}[!ht]
\mbox{\subfigure[Time domain OFDM pulse]{\includegraphics[width=2.5in]{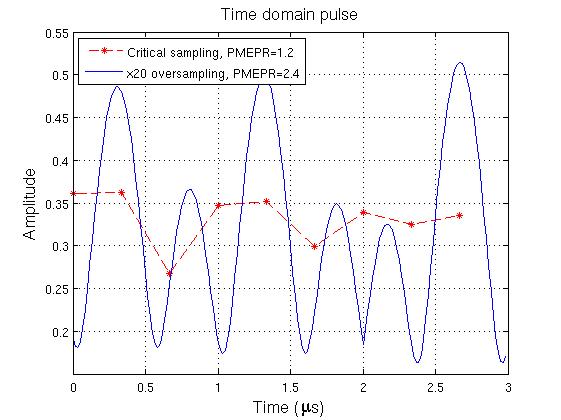}\label{fig:PMEPR_oversamp_demo}}
\subfigure[Autocorrelation output]{\includegraphics[width=2.5in]{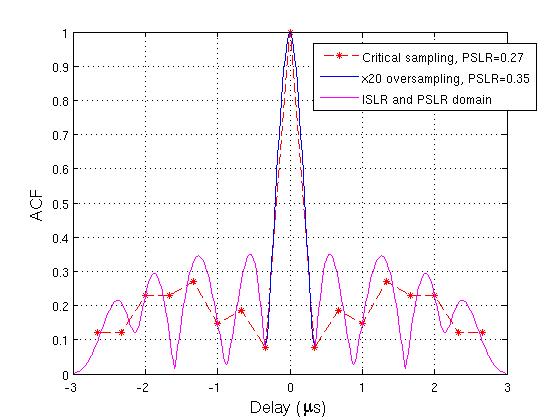}\label{fig:ACF_oversamp_demo}} }
\caption{Pulse amplitude versus time of an uncoded OFDM pulse and autocorrelation function versus range. $N=3$ and $K=3$. Both functions are plotted at the critical sampling rate $f_s=B$ and with oversampling $f_s=20B$.} \label{fig:ACF_PMEPR_demo}
\end{figure}

An important comment to make at this stage concerns the finesse of the sampling that we consider in each of our objective functions. In Eq.~\ref{eq:PSLR}, Eq.~\ref{eq:ISLR} and Eq.~\ref{eq:PMEPR}, we have assumed to work at the critical sampling rate $f_s=B$ such that the sampling period is $t_s=t_b/N$, as a result of the relationships that govern the OFDM structure. In Fig.~\ref{fig:ACF_PMEPR_demo} we stress the impact of oversampling. Because of the quick temporal variations of the OFDM signal the PMEPR will not be the same at the critical sampling rate than it will be with an oversampling factor of 20. In \cite{Lellouch13}, we stressed that the autocorrelation function of multicarrier signals presents strong variations between consecutive critically sampled instants, which is not the case of phase coded signal with a single carrier. Practically, the relevance to include or not oversampling in this analysis shall be evaluated with regard to the implementation strategy in terms of the baseband OFDM signal sampling frequency. In this work, we thus choose to oversample the OFDM signal before evaluating our objective functions. In Fig.~\ref{fig:ACF_oversamp_demo}, we show how both PSLR and ISLR calculations exclude the values around the main peak. Also discussed in \cite{Lellouch13}, the total peak extent is equal to twice the Rayleigh resolution, which in time is $2/B$.

\subsubsection{Optimization strategy}\label{sec:sect3b}
Thus, in the process of designing an OFDM pulse, extra care shall be employed so that the signal results in tolerable PMEPR as well as reasonable sidelobe level. Note however that in this statement we have assumed that the processing was based on a correlation function. In our case, this is valid when we intend to use our conventional processing. However, if rather, we decide to use our frequency domain processing, we have recalled in Table~\ref{tab:processing_charact} that both the range and Doppler sidelobes were phase codes independent. In that case, the optimization can focus on the PMEPR only. In a nutshell, this analysis indicates that with our conventional processing we are interested in optimizing 3 objectives, while with our frequency domain processing, we would have a single objective. We thus have to come up with two solutions, a multi-objective solution and a single objective one. 

\subsubsection{Parameters for optimization}\label{sec:sect3c}
We have seen in Eq.\ref{eq:OFDMbaseband} that the parameters still available after our initial design in section~\ref{sec:sect2} were the weights $w_n$ and the phase codes $a_{n,k}$. We decide to leave the weights and focus primarily on the phase codes. To justify this approach, we point out that the scattering centres of a target resonate variably at different frequencies, therefore it may be desirable to leave the weights for an optimization that would account for the target model as we will see in section~\ref{sec:sect6}. For now, we assume equal weights such that the pulse has normalized energy. The phase codes are also assumed to have constant modulus. Each $a_{n,k}$ is thus a number on the unit complex circle.

\subsubsection{Existing optimization methods}\label{sec:sect3d}
The problem of minimizing the PSLR and the PMEPR has been extensively researched since the emergence of the multicarrier concept. Levanon et al. \cite{Levanon04} have reported some the major contributions in that field. Two strategies have popped up. The first strategy assumed identical sequences, such that all subcarriers are assigned the same phase code. Optimizing the PMEPR of the pulse results in optimizing the PMEPR of a single symbol. Newman, Schroeder and Narahashi have suggested different phasing methods to decrease the PMEPR. In their concept the weights $w_n$ are complex values with quadratic dependence on $n$. PMEPR as low as 2 (3dB) can be obtained for any relevant value of $N$ (up to 65,000). The second design strategy is based on modulating all $N$ subcarriers with consecutive ordered cyclic shifts (COCS) of an ideal chirplike sequence (CLS) of length $K$. For example, a OFDM pulse based on COCS of P4 codes can give PMEPRs below 2 and PSLR below -15 dB for a large range of $N$ (between 0 and 70).

\subsubsection{Motivation for genetic algorithm optimization methods}\label{sec:sect3e} 
Although the aforementioned techniques produce excellent results in terms of PMEPR alone or PMEPR and PSLR combined, we suggest to introduce genetic algorithm based methods in this field for three main reasons. Firstly, it will diversify the solutions and increase the potential number of good codes. Not necessarily the best but those good enough for the application. Secondly, we can tune the objective functions to put emphasis on some criteria. For example, if we wish to have extremely low sidelobes close to the main peak in the autocorrelation function and can tolerate higher values further away, the objective function can be modified with no harm. Another example is the case of a banned sub-band. Assume that in the presence of a jammer we ban the use of one or few subcarriers. The energy on these subcarriers could be picked up and the presence of our transmission would be revealed. If we rely on any of the previous strategies, the loss of one or more subcarriers can potentially destroy the PMEPR, the PSLR or both. Another interesting example is the use of sparse spectrum OFDM pulses to mitigate the range ambiguity in SAR OFDM \cite{Riche12}. In both cases, the genetic algorithm based optimization can come up with sets of phase codes that will improve both the PMEPR and the PSLR in either of these configurations. Thirdly, we can also optimize the ISLR which is not in the focus of the other methods. Not only we can optimize it solely but we can optimize it together with the PSLR and the PMEPR by means of multi-objective optimization (MOO) techniques.   

\section{Optimization techniques}\label{sec:sect4}
In this section, we present the genetic algorithm based techniques that we use to find our phase codes. 

\subsubsection{Problem encoding}\label{sec:sect4a}
The first step in the implementation of any genetic algorithm is to generate an initial population. Following the canonical genetic algorithm guideline \cite{Whitley}, this implies encoding each element of the population into a binary string. Note however that techniques based on real numbers have also been developed \cite{Agrawal94}. The MOO-GA that we discuss in section~\ref{sec:sect4c} applies one of them. For now, we simply encode one phase code (value between 0 and $2\pi$) into a string of $q$ genes. When we are dealing with a pulse composed of $N$ subcarriers and $K$ symbols, we end up with $NK$ strings of $q$ genes each. Stacking these strings together we create one element of the population, which is then formed from $Q=NKq$ genes. This element is called a chromosome. If we require to use for example binary phase shift keying (BPSK), then $q=1$, quadrature phase shift keying (QPSK), then $q=2$. In the more general case where we have no restriction we can consider the largest value authorized by our system. In our case we use $q=18$. The resolution in angle is then $\Delta\theta=2\pi/2^q\simeq0.024$ mrad. Note that the larger $q$ the slower the algorithm. This is caused by the increase of the search space as discussed hereafter. With the values of $N$ and $K$ that we consider in this paper, the chromosome length can be as large as 9000, ($N=500$ and $K=1$). The search space $\mathcal{S}$ "reduces" to the binary strings of length $Q$. 

\subsubsection{Population size}\label{sec:sect4b}
To understand what the population size $L$ shall be, we followed the guideline given in \cite{Reeves03}. The starting point is to say that every point in the search space shall be reachable from the initial population by crossover only. This can happen only if there is at least one instance of every gene at each locus in the entire population. On the assumption that every gene is generated with random probability (P(1)=1/2 and P(0)=1/2) the probability that at least one gene is present at each locus is given by:
\begin{equation} \label{eq:ProbaFullSpaceSearch}  
P=(1-(1/2)^L)^{Q}
\end{equation}
With $Q=9000$ we see that $L\simeq 23$ when we take $P=99.9\%$ and $L\simeq 26$ when we take $P=99.99\%$.

\subsection{Genetic algorithm}\label{sec:sect4c} 
The genetic algorithm implemented in this work is a two-stage process. Goldberg defined this class of genetic algorithms as simple genetic algorithms (SGA) \cite{Whitley}. It starts with the current population. Then selection is applied to form the intermediate population. Next, recombination and mutation are applied to form the next population. The process of going from the current population to the next population represents one generation of the execution of the genetic algorithm. We mentioned earlier that we make use of this algorithm for the single objective optimization, where the objective is the PMEPR. The steps are:
\begin{enumerate}
\item Initialize the current population by generating $L$ chromosomes. Each chromosome has its $Q$ genes set to 0 or 1 with equal probability.  
\item For each chromosome, convert the $NK$ binary sequences into real numbers and create a set of $NK$ phase codes: $\exp(ja_{n,k})$.
\item For any of the $L$ sets, compute the oversampled complex OFDM signal, obtained as a result of applying an IDFT\footnote{If there are several symbols in the pulse the IDFT is applied on each vector of phase codes corresponding to one symbol and so on. At the end the outputs are stacked together to form the pulse} on the phase codes \cite{Lellouch08a}.
\item Evaluate the PMEPR according to Eq.~\ref{eq:PMEPR} and attach this number to the corresponding chromosome. 
\item Form the intermediate population by discarding the weakest element (highest PMEPR) and duplicating the strongest element (lowest PMEPR). 
\item Prepare for recombination by associating chromosomes by pair. Each pair shall be composed of two distinct chromosomes.
\item For each pair, apply a one point crossover \cite{Whitley}. This process results in the generation of $L$ offspings.
\item Apply mutation every two generations. When the generation number is odd, select at random $L_{\text{mut}}$ offsprings and for each of them apply mutation on one of their gene, again, selected at random.
\item Feed the intermediate population into the current population. 
\end{enumerate}
Steps 2 to 9 are carried out until the stopping criteria is met. The latter is characterized by two elements. A threshold on the population mean fitness that guarantees satisfactory solutions in the current population as well as a threshold on the standard deviation which assesses the convergence of the entire population towards a minimum solution.

\subsection{Multi-objective genetic algorithm}\label{sec:sect4d}
Our multi-objective problem could be solved as a single optimization problem by formulating an objective function in the form $s = \alpha PMEPR+\beta PSLR+\gamma ISLR$ and then optimizing with respect to $s$ with our GA. The drawback of that solution is judicious selection of the weights. If multiple solutions are required, the problem has to be run repeatedly for different sets of weights. To overcome this difficulty, many multi-objective evolutionary optimization algorithms have been developed, which produce a set of non dominated solutions in a single run. In this work, we make use of the well-known non-dominated sorting genetic algorithm NSGA-II \cite{Deb02}. The principal breakthrough of this algorithm is the convergence towards the true Pareto optimal set with a good spread or diversity of the solutions \cite{Seshadri06}. When selecting the best elements not only the respective fitness functions are evaluated but also the crowding distance, which tells whether this element is in a high density zone or conversely in an low density zone. At equal fronts, we select the isolated element in order to maintain diversity in the solutions. In light of this preparation we describe the steps of this algorithm, which we use with either two or three of our objective functions. In comparison to our GA, the NSGA-II uses real numbers throughout. Genetic operations such as cross-over and mutation employ a method that simulates the equivalent binary processes. A thorough review of this technique is described in \cite{Agrawal94}. 

\begin{enumerate}
\item Initialize the population with $L$ sets of $NK$ phases, all real numbers between 0 and $2\pi$ and calculate the phase codes $\exp(ja_{n,k})$.
\item For any of the $L$ sets, compute the oversampled complex OFDM signal, obtained as a result of applying an IDFT on the phase codes and calculate the autocorrelation function.
\item Evaluate the objective functions following Eq.~\ref{eq:PSLR}, Eq.~\ref{eq:ISLR} and Eq.~\ref{eq:PMEPR}.
\item Sort the chromosomes according to non-dominated sort and form the fronts ($\equiv$ rank). 
\item Evaluate the crowding distance of each chromosome.
\item Select the parent chromosomes using binary tournament selection. In binary tournaments, two chromosomes are randomly chosen and the strongest in terms of rank is selected to be in the parent population. If individuals with the same rank are encountered, their crowding distance is compared. A lower rank and higher crowding distance is the selection criteria. The parent population has a size of $L/2$.
\item Apply genetic operations such as cross-over and mutation on the selected chromosomes to produce the off-springs.
\item Combine the off-spring population with the parent population and select the best $L$ chromosomes for the next generation, again based on the rank and if needed on the crowding distance.
\end{enumerate} 
Steps 2 to 8 are carried out until the stopping criteria is met, which occurs in this case when the total number of generations has been covered.         
  
\section{Simulation results}\label{sec:sect5}
 
In this section we present our simulation results. We start with the results obtained with our GA based algorithm.
 
\subsection{Single objective: PMEPR}\label{sec:sect5a} 

\begin{figure}[!ht]
\mbox{\subfigure[Full band]{\includegraphics[width=2.5in]{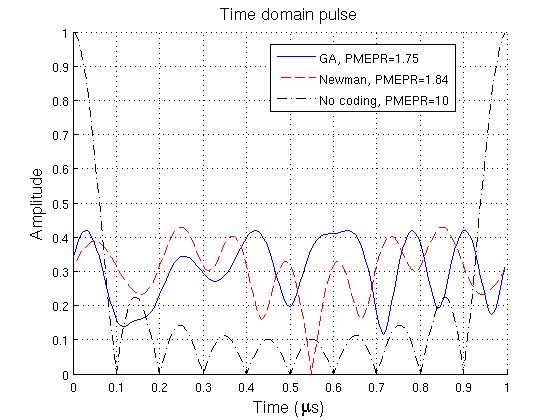}\label{fig:PMEPR_N_10_sparse_1}}
\subfigure[80$\%$ of the band]{\includegraphics[width=2.5in]{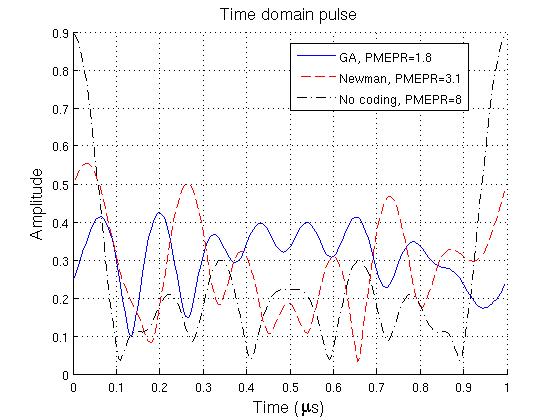}\label{fig:PMEPR_N_10_sparse_0_8}} }
\caption{Time domain signal of the OFDM symbol for different coding. When the coverage of the band is sparse our GA solution outperforms significantly the Newman phase coding.} \label{fig:PMEPR_N_10}
\end{figure}

In our simulations, we considered a population of $L=22$ chromosomes, and the number of chromosomes selected for mutation every two generations was $L_{\text{mut}}=5$. Despite the simplicity of our genetic algorithm, we are able to retrieve phase code sequences with optimal PMEPR properties. Following our previous comment, we address the interesting case of sparse spectrum. In Fig.~\ref{fig:PMEPR_N_10}, we compare the PMEPR of three sets of phase codes for two different configurations. In the first configuration, all subcarriers are enabled. We see in Fig.~\ref{fig:PMEPR_N_10_sparse_1} that our GA solution with 10 subcarriers outperforms the Newman solution. We also show the relative gain as compared to the uncoded case. The latter results in the highest PMEPR. Possibly not harmful with 10 subcarriers it cannot be tolerated with 500 subcarriers. In Fig.~\ref{fig:PMEPR_N_10_sparse_0_8} we present our GA solution when we disabled two subcarriers and compare now the PMEPR of this solution with the Newman and uncoded cases when the same subcarriers have been disabled. Our GA solution outperforms the Newman solution.
      
In Table~\ref{table:PMEPR_results_N_100}, we evaluated the PMEPR for higher numbers of subcarriers. We chose to use $N=500$ and $N=100$ to comply with our design parameters summarized in Table~\ref{tab:scenar_charact}. We observe that when we face a sparse spectrum we have no difficulty to find solutions that outperform the Newman phases. The results given in the last column were produced while using only two genes per phase, as with QPSK. This aspect is attractive in particular if we were to use a communication system for radar with a predefined alphabet, such as QPSK, etc. 
\begin{table}[!ht]
\caption{Simulation results for the PMEPR (in dB) with different levels of sparsity} 
\centering 
\begin{tabular}{l | ll | cc | cc | cc | cc } 
& && No coding && Newman && GA solution && GA solution (QPSK) &\\ [0.5ex]
\cline{1-11}  
\multirow{3}{*}{\textbf{N=100}} & Full && 100 && 1.8 && 3.3 && 3.4 &\\ 
\cline{2-11}
&70$\%$ && 70 && 4.2 && 2.9 && 3.1 &\\
\cline{2-11}
&50$\%$ && 50 && 4.1 && 3.2 && 3.3 &\\
\cline{1-11}
\multirow{3}{*}{\textbf{N=500}} & Full && 500 && 1.8 && 3.9 && 4.6 &\\ 
\cline{2-11}
&70$\%$ && 350 && 4.9 && 3.9 && 4.6 &\\
\cline{2-11}
&50$\%$ && 250 && 5.0 && 4.5 && 4.5 &\\
\end{tabular}
\label{table:PMEPR_results_N_100} 
\end{table}

In Fig.~\ref{fig:PMEPR_N_10}, we show the convergence of our GA for both cases of "no modulation" and QPSK modulation. We observe what we could intuitively guess. Because of the shorter chromosome size, the QPSK tends to converge in a more chaotic manner than our "no modulation" case. In the latter, each phase code is represented by a binary string composed of 18 bits. As a result, from one generation to the next, the genetic operations will transform the phase code set (equivalent chromosome) in a less radical fashion than with QPSK. The convergence appears thus smoother.   
\begin{figure}[!ht]
\mbox{\subfigure[No modulation (18 genes)]{\includegraphics[width=2.5in]{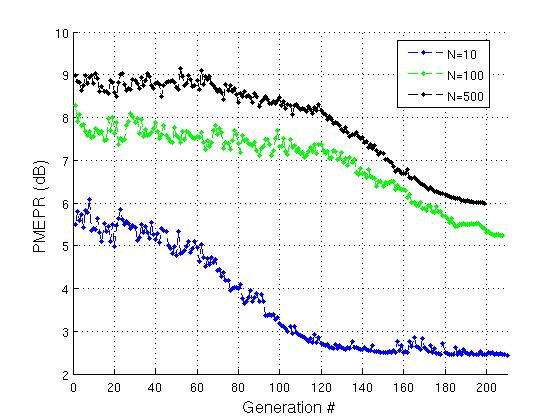}\label{fig:convergence_PMEPR_genes18}}
\subfigure[QPSK (2 genes)]{\includegraphics[width=2.5in]{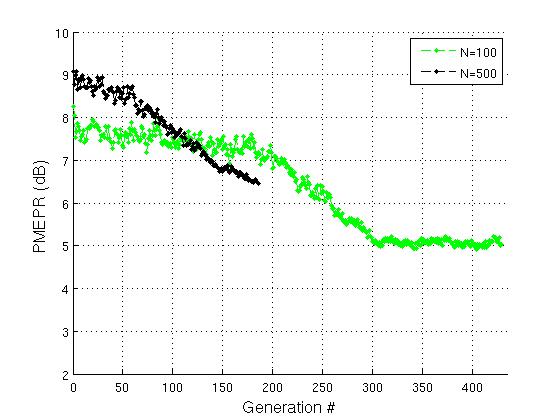}\label{fig:convergence_PMEPR_genes2}} }
\caption{Convergence of the SGA when using either 18 genes or 2 genes to describe one phase. The population contains 22 elements.} \label{fig:PMEPR_N_10}
\end{figure}

\subsection{Multiple obectives: PMEPR and PSLR}\label{sec:sect5b} 

\begin{figure}[!ht]
\mbox{\subfigure[$N$=25, $K$=4]{\includegraphics[width=2.5in]{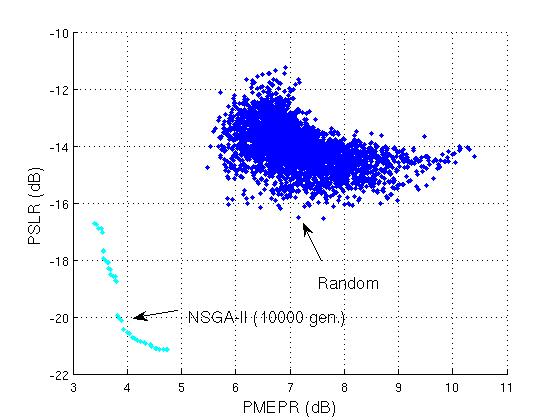}\label{fig:Comparison_after_1_and_10000_runs_N25_K4}}
\subfigure[$N$=125, $K$=4]{\includegraphics[width=2.5in]{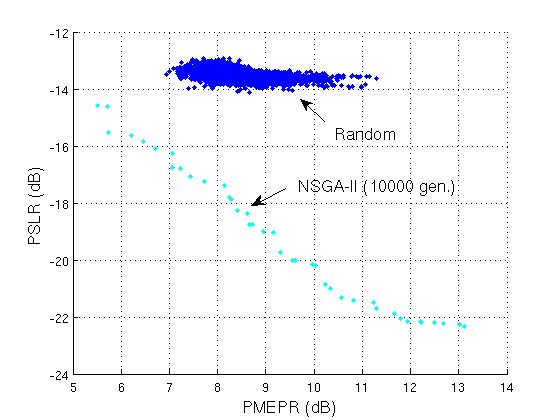}\label{fig:Comparison_after_1_and_10000_runs_N125_K4}} }
\caption{Comparison between random phases and phases resulting from the NSGA-II optimization.} \label{fig:Compa_run1_run10000}
\end{figure}

One of the benefit of the NSGA-II as compared to the prior version, NSGA is its fast computing time as a result of the selection strategy based on chromosome rank and crowding distance. On top, the use of binary simulated crossover and mutation improves the algorithm complexity. The large chromosome sizes from our GA are simply suppressed as we work throughout the algorithm with sets of real numbers for the phase codes. In Figs.~\ref{fig:Compa_run1_run10000}, we present the improvement obtained with the NSGA-II when we intend to find a set of phase codes for our two design cases given in Table~\ref{tab:scenar_charact}. In both cases, we have taken a smaller number of subcarriers, but the time bandwidth product remains the same, $NK=100$ and $NK=500$ respectively. For comparison, we plotted 100 realizations of the initial random population. With a population size equal to $L=40$, our clouds comprise 4000 points. After 10000 generations our set of optimal solutions is considerably improved.

\begin{figure}[!ht]
\mbox{\subfigure[$N$=25, $K$=4]{\includegraphics[width=2.5in]{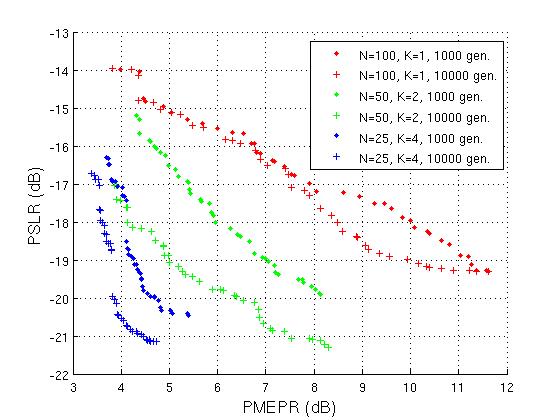}\label{fig:Comparison_after_1000_and_10000_runs_NK_100}}
\subfigure[$N$=125, $K$=4]{\includegraphics[width=2.5in]{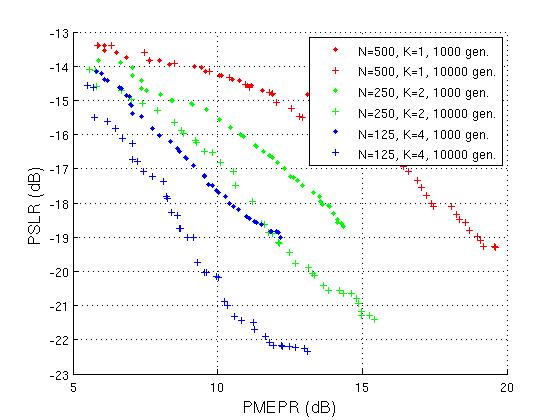}\label{fig:Comparison_after_1000_and_10000_runs_NK_500}} }
\caption{Convergence of the NSGA-II optimization.} \label{fig:Compa_3runs}
\end{figure}

In Fig.~\ref{fig:Compa_3runs}, we propose to compare three cases, all having equal time bandwidth product, like in our design. We see that when the same number of generations is considered, the design with the smallest number of subcarrier will provide the best pareto front. Again, this results confirms an intuitive guess.

\subsection{Multiple objectives: PMEPR, PSLR and ISLR}\label{sec:sect5c} 

\begin{figure}[!ht]
\centering
\includegraphics[scale=0.4]{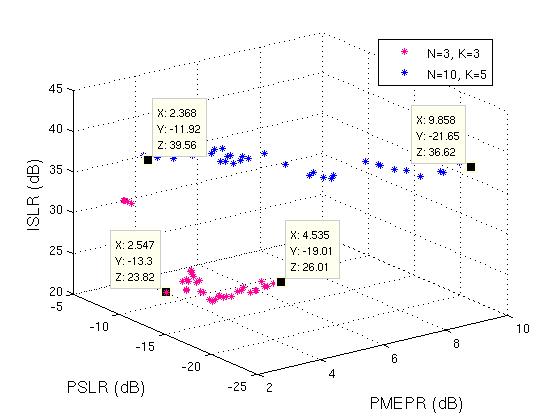} 
\caption{Pareto front after 10000 generations for two pulse design.} 
\label{fig:3Dpareto}
\end{figure}

As opposed to the single optimization case, which we presented in \cite{Lellouch13}, we can feed our three objective functions of interest into our MOO-GA to provide us with a 3D map of solutions, related to a particular design. In light of the intended application and the design constraints the most suitable solution can be selected. Fig.~\ref{fig:3Dpareto} shows the 3D pareto fronts of the design cases evaluated in \cite{Lellouch13}. 

\section{Case study: GA for PMEPR optimization in a target detection enhancement procedure}\label{sec:sect6}
\begin{figure}[!ht]
\centering
\includegraphics[scale=0.4]{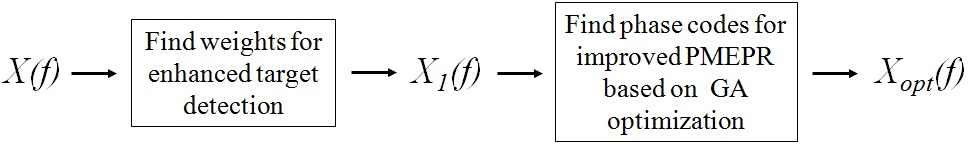} 
\caption{OFDM pulse optimization in two steps.} 
\label{fig:OFDM_opti_2steps}
\end{figure}

In this section we show how our single objective GA optimization can be integrated in a procedure intended to enhance the target detection. Our OFDM pulse is tailored in two steps. Firstly, we find the best weights to enhance the detection. Secondly, we run our GA to find satisfactory sets of phase codes. Our procedure is presented in Fig.~\ref{fig:OFDM_opti_2steps}

\subsection{SNR as the design metric}\label{sec:sect6a}
Unlike traditional notions where imaging-based metrics for waveform design are best for target detection and classification, which lead to our objectives of low PSLR and low ISLR, another approach relies on the well-known SNR metric. In that case, the optimum waveform will not necessarily have good PSLR nor ISLR properties.

\subsection{Transmitted signal receiver-filter pair}\label{sec:sect6b}
It has been shown \cite{Wilkinson98} that the signal at the output of the matched filter $V_s(f)$, in a radar receiver, is given by the product of the target reflectivity spectrum $\varsigma(f)$, the waveform spectrum $X(f)$ and the filter transfer function $H(f)$:
\begin{equation} \label{eq:vBB}
V_{s}(f) =\varsigma(f+f_c)X(f)H(f).
\end{equation}

\begin{figure}[!ht]
\centering
\includegraphics[scale=0.4]{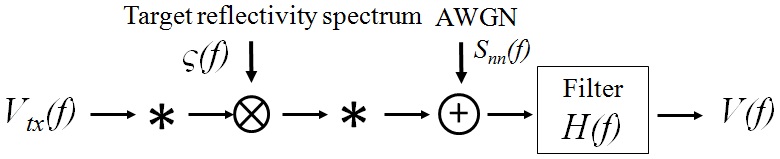} 
\caption{Model of the received signal in the frequency domain.} 
\label{fig:Signal_transformation_freq}
\end{figure}

The frequency domain operations undergone by the waveform are sumarized in Fig.~\ref{fig:Signal_transformation_freq}, where $V_{tx}(f)$ is the Fourier transform of the transmitted signal $v_{tx}(t)$:

\begin{equation} \label{eq:vtx} 
v_{tx}(t) = x(t)\exp(j2\pi f_ct),
\end{equation}
$f_c$ is the carrier frequency and $x(t)$ is the OFDM pulse as given in Eq.~\ref{eq:OFDMbaseband}. In this analysis, we consider that the pulse is composed of a single OFDM symbol.

\subsubsection{Optimum receiver-filter pair}\label{sec:sect6b1}
In the case of an extended target, unlike the standard matched filtering problem applicable when the target is modelled as a point scatterer, the optimal receiver filter shall be matched to the waveform scattered by the target, not the transmitted target itself \cite{Bell93}. When this is the case, the maximum SNR at time $t_0$ is given by the following equation \cite{Bell93}:
\begin{equation} \label{eq:SNR_opt} 
\left(\frac{S}{N}\right)_{t_0} = \int_{-\infty}^\infty \frac{|\varsigma(f+f_c)X(f)|^2}{S_{nn}(f)}df.
\end{equation}

While previous works have researched solutions of this SNR optimization problem in the time domain, to find appropriate $x(t)$, we argue that we can benefit from the OFDM structure and find directly $X(f)$. Assuming our OFDM signal to be zero outside its bandwidth\footnote{The OFDM spectrum of the baseband pulse is considered to spread between 0 and $B$ rather -$B$/2 to $B$/2}, and considering that the stationary additive Gaussian noise is white, with one-sided power spectral density $S_{nn}(f)=N_0$, we can modify Eq.~\ref{eq:SNR_opt} into: 
\begin{equation} \label{eq:SNR_opt_bb} 
\left(\frac{S}{N}\right)_{t_0} = \frac{1}{N_0}\int_0^B |X(f)\varsigma(f+f_c)|^2df,
\end{equation} 
If we know the target reflectivity spectrum, we can rewrite Eq.~\ref{eq:SNR_opt_bb} in discrete terms:
\begin{equation} \label{eq:SNR_opt_bb_discrete} 
\left(\frac{S}{N}\right)_{t_0} = \frac{1}{N_0}\sum_{n=0}^{N-1} |X[n]|^2 \cdot |\varsigma[n]|^2,
\end{equation}  
where $X[n]=\frac{w_n a_n}{\sqrt{N}}$ is our discrete OFDM spectrum, $X[n]=X(n\Delta f)$ and $n$ takes values from 0 to $N-1$. Following this expression, when weights are equal to 1 the signal is normalized (unit energy). Assuming that we transmit at all subcarriers, our optimization problem reduces in finding the weights $w_n$ such that:
  
\begin{equation} \label{eq:OFDMsymbolfreq}  
\text{arg }\underset{w_n}{\operatorname{max}}  \sum_{n=0}^{N-1} w_n^2\cdot |\varsigma[n]|^2, \text{ s.t. } \left\{
    \begin{array}{ll}
        w_n\neq 0 \\
        \sum_{n=0}^{N-1} w_n^2=N
    \end{array} \right.
\end{equation}

A solution to this problem is obtained when $w_n\propto |\varsigma[n]|$.

\subsection{Simulation setup and results}\label{sec:sect6c}
In our analysis we choose to work at X-band. We consider a 2 GHz bandwidth $B$ centred around 10 GHz, $f_c$ = 9 GHz. Before we elaborate on the reflectivity spectrum of our target within this band we stress the need to normalize $\varsigma[n]$ and describe our normalization strategy. 

\subsubsection{Target reflectivity spectrum normalization}\label{sec:sect6c1}
For relevant comparisons, we normalize the discrete target reflectivity spectrum following the strategy suggested in chapter 14 in \cite{Gini12}. Essentially, when a flat spectrum OFDM pulse of unit energy interacts with the normalized reflectivity spectrum $\varsigma_{\text{norm}}[n]$, the frequency-domain reflected signal shall have unit average power. The discrete elements of our flat spectrum OFDM pulse of unit energy are given by $X[n]=a_n/\sqrt{N}$. Hence, we find for $\varsigma_{\text{norm}}[n]$:
\begin{equation} \label{eq:H_norm} 
|\varsigma_{\text{norm}}[n]|^2 = \frac{N^2}{\sum_{n=0}^{N-1} |\varsigma[n]|^2}\cdot |\varsigma[n]|^2,
\end{equation}  
 
\subsection{Complex target model}\label{sec:sect6c2}
\begin{figure}[!ht]
\mbox{\subfigure[Synthetic target]{\includegraphics[width=2.5in]{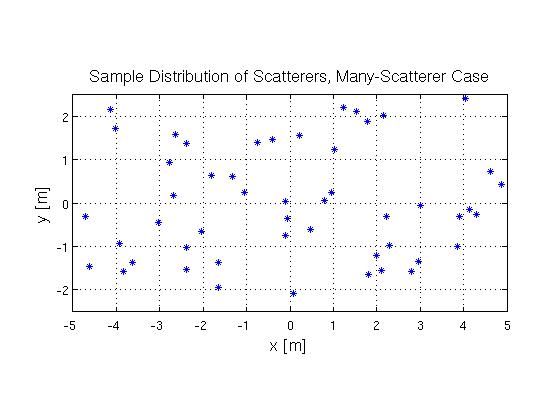}\label{fig:Synthetic_target}}
\subfigure[Reflectivity power spectrum]{\includegraphics[width=2.5in]{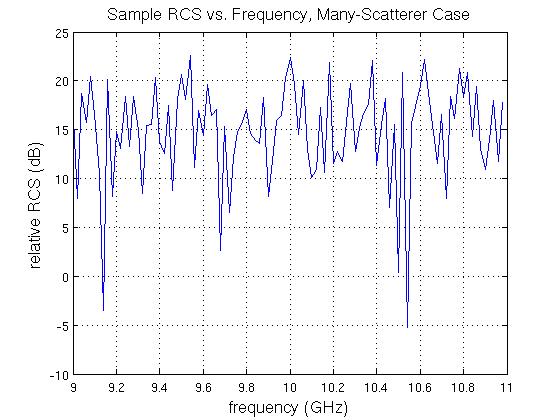}\label{fig:Synthetic_target_RCS}} }
\caption{Complex target made of $P=$ 50 point scatterers.} \label{fig:complex_target_model}
\end{figure}

In our simulation we construct a synthetic target from $P=50$ point scatterers with equal unit reflectivity $\varsigma_i=\sqrt{\sigma_i}=1$ and located at ranges $R_i$ from the radar. The individual point scatterers are assumed to be perfectly conducting spheres, large enough to have a reflectivity constant within the band of interest. It can be shown \cite{Richards10} that the compound target reflectivity spectrum $\varsigma(f)$ is equal to: 
\begin{equation} \label{eq:reflectivity_complex_freq} 
\varsigma(f) = \sum_{i=1}^P\sqrt{\sigma_i}\exp(-j4\pi f\frac{R_i}{c}).
\end{equation}
The point scatterers are randomly distributed within a rectangle, 5 meters wide and 10 meters long, whose center is 10 km away from the radar along the x axis. Fig.~\ref{fig:Synthetic_target} shows the position of the scatterers, while Fig.~\ref{fig:Synthetic_target_RCS} gives the target reflectivity power spectrum. As expected, we observe strong variations of the power spectrum within the frequency band.  

\section{SNR and PMEPR improvements}\label{sec:sect6c3}
Following the above methodology, in the first step, the $N$ weights are derived from the normalized target reflectivity spectrum, so that the intermediate OFDM spectrum $X_1(f)$ has enhanced detection capability. In the second step, $X_1(f)$ is fed into our GA optimization to find a set of $N$ phase codes that will improve the PMEPR. In the end, as shown in Fig~\ref{fig:OFDM_opti_2steps} our pulse $X_{opt}(f)$ has enhanced detection capabilities for the target of interest and a reasonably low PMEPR. 

\begin{figure}[!ht]
\mbox{\subfigure[Weights]{\includegraphics[width=2.5in]{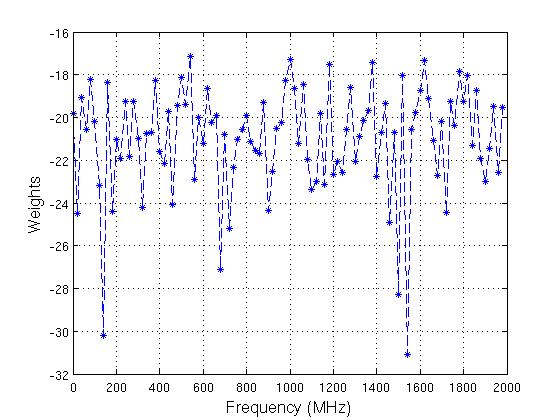}\label{fig:case_study_weights}}
\subfigure[PMEPR improvement]{\includegraphics[width=2.5in]{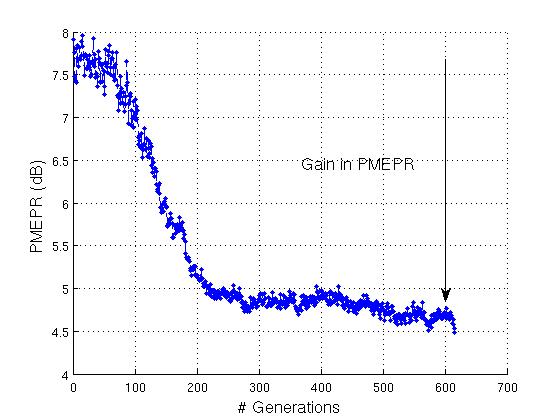}\label{fig:case_study_PMEPR}} }
\caption{Use of the single objective GA to optimize the PMEPR based when an optimized set of weights is applied on the subcarriers.} \label{fig:complex_target_model}
\end{figure}  
In table~\ref{tab:result_enhancement}, we show the gain in dB of the weights optimization method.

\begin{table}[h]
\caption{Result for the detection enhancement} 
\centering 
\begin{tabular}{c | c | c} 
  & Flat spectrum & Optimal weights
\\ [0.5ex]
\hline 
Average power (dB)& 0 & 2.4 \\ 
\hline 
\end{tabular}
\label{tab:result_enhancement}
\end{table}
 
\section{Conclusion}\label{sec:sect7}
In this paper, we showed that GA based techniques are suitable to optimize or improve the design of OFDM pulses for radar. We inspected the possibility to incorporate these optimization techniques in the more general waveform design in regards of two processing solutions. In our first conventional processing solution, primarily, the PSLR and the PMEPR need to be optimized. The former is a figure-of-merit for the detectability of small targets and the latter relates to the severity of the distortion. The MOO-GA based algorithm NGSA-II has been used to optimize this multi-objective problem. We observed that with the PMEPR and the PSLR for the objective functions a substantial improvement is achieved as compared to the case where random coding is applied. We also showed that the more symbols in the pulse the better the optimization in PMEPR and PSLR, for the same time bandwidth product. In our second, frequency domain processing solution, we stressed that the main focus should be on the minimization of the PMEPR. Therefore, the single objective GA can be used. We showed that this evolutionary technique can produce solutions that outperform the most robust methods. We also demonstrated that in some relevant cases, when the OFDM spectrum is sparse, our GA solution gives outstanding results. Finally, we presented a case study to attest the relevance of our GA optimization method where the optimization relies on the set of phase codes of the OFDM pulse. In this case study, we showed that the problem of finding a suitable radar pulse for enhancing the detection of a known target could be solved in two steps. In the first step, the weights of the subcarriers are selected in light of the target reflectivity spectrum. In the second step, our GA based optimization method is applied on this set of weights to search for the most appropriate phase codes that would minimize the PMEPR. For completeness we may conclude saying that when both PSLR and ISLR need to be optimized, we suggest to use a MOO-GA based on two objectives rather than three, and use the PMEPR as a constraint.

\end{document}